\documentclass[sigconf]{acmart}

\usepackage{graphicx}
\usepackage{amsmath}
\usepackage{multirow}
\usepackage[ruled,vlined]{algorithm2e}

\AtBeginDocument{%
  }

\copyrightyear{2026}
\acmYear{2026}
\setcopyright{cc}
\setcctype{by}
\acmConference[Mulsemedia '26]{2nd International Workshop on Multi-Sensorial Media and Applications}{November 10--14, 2026}{Rio de Janeiro, Brazil}
\acmBooktitle{2nd International Workshop on Multi-Sensorial Media and Applications (Mulsemedia '26), November 10--14, 2026, Rio de Janeiro, Brazil}
\acmDOI{10.1145/3841456.3841509}
\acmISBN{979-8-4007-2945-4/2026/11}

\begin{document}

\title{RegCL: Compact Continual SAM Adaptation for Visual Grounding in Multi-Sensorial Media}

\author{Yuan-chen Shu}
\orcid{0009-0005-7318-7720}
\affiliation{%
  \department{Wangxuan Institute of Computer Technology}
  \institution{Peking University}
  \city{Beijing}
  \country{China}
}
\email{andrewsu317@stu.pku.edu.cn}

\author{Zhiwei Lin}
\orcid{0000-0002-8841-7673}
\affiliation{%
  \department{Wangxuan Institute of Computer Technology}
  \institution{Peking University}
  \city{Beijing}
  \country{China}
}
\email{zwlin@pku.edu.cn}

\author{Xiaoyu Zhou}
\orcid{0000-0002-6562-5752}
\affiliation{%
  \department{Wangxuan Institute of Computer Technology}
  \institution{Peking University}
  \city{Beijing}
  \country{China}
}
\email{xiaoyu\_zhou@stu.pku.edu.cn}

\author{Yongtao Wang}
\correspondingauthor
\orcid{0000-0003-1379-2206}
\affiliation{%
  \department{Wangxuan Institute of Computer Technology}
  \institution{Peking University}
  \city{Beijing}
  \country{China}
}
\email{wyt@pku.edu.cn}


\begin{abstract}
% Continual Learning (CL), the ability to acquire new knowledge without forgetting previous information, is crucial for widely used neural networks models.
% 
% Simply updating a neural network's parameters for new tasks often leads to Catastrophic Forgetting, where performance on earlier tasks degrades.
% 
% Previous methods address this by either restricting parameter updates, replaying past data, or adding specialized modules to separate new and existing knowledge.
% However, these methods do not perform well on foundation models, such as the segment anything model (SAM).
% 
% To overcome these limitations, we propose RegCL, a novel non-replay CL framework that integrates multi-domain knowledge continuously through model merging.
% 
% We demonstrate its effectiveness on SAM by merging the parameters of its adaptation modules (e.g., LoRA modules) trained on different domains.
%
% The merging process is guided by weight optimization, minimizing prediction discrepancies between the merged model and each domain-specific models.
%
% Experiments show that RegCL delivers strong continual learning performance across diverse datasets, effectively consolidating cross-domain knowledge.

Multi-sensorial media systems, including AR/VR, remote operation, and embodied AI, require visual grounding modules that remain reliable as sensing environments and application domains evolve.
The Segment Anything Model (SAM) provides a strong foundation for dense visual segmentation, but its performance degrades on specialized and dynamically arriving domains such as medical imagery, camouflaged scenes, and shadow-dominant environments.
Existing continual learning methods often rely on replay data or growing domain-specific modules, limiting compact deployment in evolving media pipelines.
To address this issue, we propose RegCL, a non-replay continual adaptation framework that consolidates multi-domain segmentation knowledge into a single SAM adapter through incremental model merging.
RegCL merges lightweight adaptation modules, e.g., LoRA-style AugModules, by optimizing prediction consistency between the merged model and domain-specific adapters while carrying forward compact historical feature statistics.
Experiments across five heterogeneous segmentation datasets show that RegCL achieves strong retention and adaptation under domain-incremental learning, outperforming competitive non-replay continual learning and merging baselines.
These results suggest that RegCL can serve as a compact visual adaptation component for evolving multi-sensorial media pipelines.
The code is available at \href{https://github.com/Anderw-S/RegCL}{https://github.com/Anderw-S/RegCL}

\end{abstract}




% \received{11 August 2026}
% \received[revised]{11 August 2026}
% \received[accepted]{11 August 2026}


\maketitle

\section{Introduction}
\label{sec:intro}

% Foundation models have become an increasingly important paradigm for visual recognition and dense prediction.
% In particular, the Segment Anything Model (SAM)~\cite{SAM} has demonstrated remarkable zero-shot segmentation capability on natural images, making it an attractive starting point for downstream segmentation problems.
% However, a growing body of evidence shows that SAM does not transfer equally well to specialized domains such as medical images, camouflaged scenes, and shadow-dominant environments, where substantial domain shift often leads to unsatisfactory segmentation quality~\cite{SAM-Adapter,Sam-adapter-med}.
% This gap has motivated a line of work on adapting SAM to domain-specific segmentation tasks.
Multi-sensorial media systems, including AR/VR, remote operation, and embodied AI, depend on visual grounding modules that can convert changing visual observations into object- or region-level representations for interaction and downstream sensory feedback~\cite{RobustVisH,HapticCompression}.
Foundation models have become an increasingly important paradigm for this visual analysis layer because they provide transferable representations for recognition and dense prediction.
In particular, the Segment Anything Model (SAM)~\cite{SAM} has demonstrated remarkable zero-shot segmentation capability on natural images, making it an attractive starting point for segmentation-driven media analysis.
However, a growing body of evidence shows that SAM does not transfer equally well to specialized domains such as medical images, camouflaged scenes, and shadow-dominant environments, where substantial domain shift often leads to unsatisfactory segmentation quality~\cite{SAM-Adapter,Sam-adapter-med}.
This gap motivates continual adaptation methods that keep the visual grounding component reliable as new application domains emerge.

A practical solution is parameter-efficient fine-tuning (PEFT), which adapts a frozen pre-trained model through lightweight modules such as adapters or low-rank updates~\cite{houlsby2019parameter,lora,adaptformer}.
For SAM, representative methods such as SAM-Adapter~\cite{SAM-Adapter} and Medical SAM Adapter~\cite{Sam-adapter-med} show that updating only a small fraction of parameters can substantially improve performance on challenging segmentation benchmarks.
Nevertheless, these approaches are predominantly developed under a \emph{one-step} adaptation setting: each model is tuned for one target domain in isolation, and the resulting parameters are typically specialized to that single domain.
When a new domain arrives, the common practice is either to fine-tune again from the base model or to maintain an additional domain-specific module, neither of which provides a satisfactory solution for long-term, evolving deployments.

% In realistic applications, however, target domains rarely arrive all at once.
% Instead, new domains emerge continuously, and the model is required to incorporate new knowledge without sacrificing previously acquired capabilities.
% This naturally leads to the continual learning (CL) setting.
% A direct sequential fine-tuning pipeline is inadequate because adapting to a new domain often overwrites previously learned knowledge, leading to catastrophic forgetting~\cite{de2021continual,EWC}.
% Classical CL methods address this issue through regularization~\cite{EWC}, replay~\cite{icarl}, or architecture expansion~\cite{rusu2016progressive}.
% While effective in conventional settings, these strategies are less suitable for adapter-based adaptation of foundation models for segmentation: replay requires storing historical data, regularization may restrict the plasticity of lightweight adaptation modules, and architecture-based solutions usually preserve old knowledge by maintaining task-specific parameters whose storage and management cost grows with the number of domains.
In realistic media deployments, target domains rarely arrive all at once.
Instead, sensing environments, object appearances, and application contexts evolve over time, and the visual grounding model must incorporate new knowledge without sacrificing previously acquired capabilities.
This naturally leads to the continual learning (CL) setting.
A direct sequential fine-tuning pipeline is inadequate because adapting to a new domain often overwrites previously learned knowledge, leading to catastrophic forgetting~\cite{de2021continual,EWC}.
Classical CL methods address this issue through regularization~\cite{EWC}, replay~\cite{icarl}, or architecture expansion~\cite{rusu2016progressive}.
While effective in conventional settings, these strategies are less suitable for adapter-based adaptation of foundation models for segmentation: replay requires storing historical data, regularization may restrict the plasticity of lightweight adaptation modules, and architecture-based solutions usually preserve old knowledge by maintaining task-specific parameters whose storage and management cost grows with the number of domains.

Recent efforts have started to study SAM under dynamic-domain continual adaptation.
MoDA~\cite{moda} introduces a continual adaptation benchmark for SAM and proposes MoDA, a mixture-of-domain-adapters strategy for preserving domain-specific information during sequential learning.
SAMCL~\cite{samcl} further improves continual SAM adaptation by learning separate AugModules together with a Module Selector for inference-time routing.
Although effective, these methods still rely on domain-specific module libraries and selection mechanisms, so the overall adaptation state continues to expand as new domains accumulate.
This limits scalability and departs from the appealing goal of maintaining a compact, unified model throughout continual adaptation.

Model merging offers a different perspective on this problem.
Instead of constraining parameter updates during training, it combines multiple task-specific solutions directly in weight space~\cite{Soups,TA,finshermerging,Regmean,Ties,dare}.
For adapter-based SAM adaptation, this is especially appealing: each domain can first be adapted independently, and then consolidated into a running adapter without revisiting previous data.
However, simple weight-space interpolation is often unstable because independently trained updates may interfere with one another, especially when different tasks induce conflicting parameter changes~\cite{Ties,dare}.
Moreover, most merging methods assume simultaneous access to all models rather than incremental, memory-limited consolidation~\cite{magmax,DOP}.

% To address these challenges, we propose \textbf{RegCL}, a novel non-replay continual learning framework for adapting SAM through \emph{incremental model merging}.
% Rather than preserving old knowledge by storing past data or maintaining a growing collection of domain-specific modules, RegCL continuously integrates newly learned domain knowledge into a compact adapter via merging.
% Specifically, it merges parameters of trained adaptation modules (e.g., LoRA modules) by optimizing prediction consistency between the merged model and new domain-specific models.
% Meanwhile, RegCL carries forward compact historical statistics, enabling efficient incremental consolidation without replaying previous task data.
% In this way, RegCL preserves the parameter-efficiency advantages of PEFT while avoiding the storage growth of module-library-based continual adaptation.

\begin{figure}[t]
    \centering
    \includegraphics[width=0.7\linewidth]{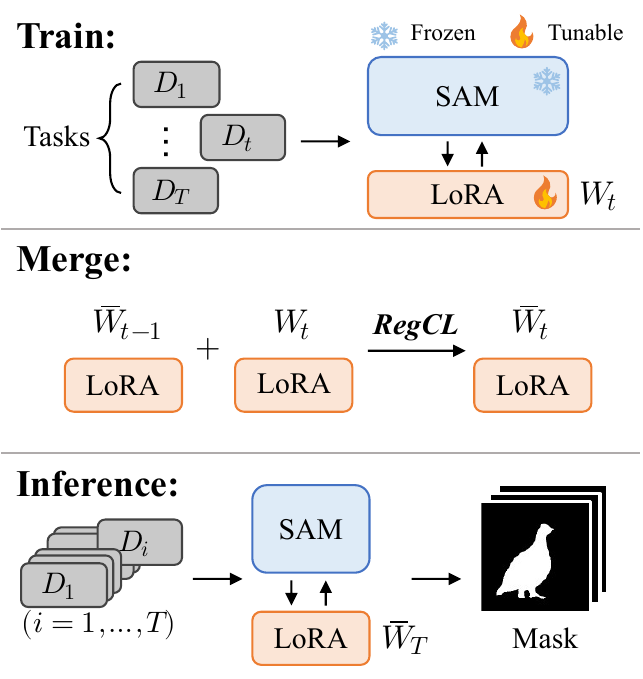}
    \caption{
        \textbf{Illustration of RegCL for continual learning.}
        RegCL merges weights from independent fine-tuned models in a continual learning setting.
    }
    \label{fig:intro}
\end{figure}
To address these challenges, we propose \textbf{RegCL}, a non-replay continual learning framework that keeps SAM compact while adapting to evolving visual domains.
As shown in Figure~\ref{fig:intro}, RegCL can be viewed through three stages: training, merging, and inference.
When a new domain arrives, RegCL trains a lightweight task-specific adapter and then consolidates it with the previously merged adapter.
Rather than replaying data or expanding module library, RegCL retains a single adapter for inference together with compact feature statistics for future consolidation.
The merge preserves the functional responses of task-specific adapters on corresponding feature distributions, enabling compact continual consolidation for deployable multi-sensorial media pipelines.
% This design preserves the parameter-efficiency advantages of PEFT and avoids the storage growth of module-library-based continual adaptation, making RegCL suitable for deployable multi-sensorial media pipelines.

% We evaluate RegCL on multiple segmentation datasets spanning medical, camouflaged, and shadow object segmentation.
% Extensive experiments show that RegCL achieves a strong balance between adaptation and retention, outperforming competitive continual learning baselines while maintaining a constant model size across tasks.
We evaluate RegCL on multiple segmentation datasets spanning medical imagery, camouflaged objects, and shadow-dominant scenes.
RegCL achieves a strong balance between adaptation and retention, outperforming competitive continual learning baselines while maintaining a constant model size.

The main contributions of this paper are summarized as follows:
\begin{itemize}
    % \item We propose RegCL, a non-replay continual learning framework for adapter modules that continuously adapts SAM to new domains through incremental model merging.
    \item We formulate compact continual SAM adaptation as a visual grounding problem for evolving multi-sensorial media pipelines.

    % \item We develop an efficient merging strategy for lightweight adaptation modules, which consolidates domain knowledge without requiring access to previous task data.
    \item We develop RegCL, an efficient incremental merging strategy on adapters that consolidates domain knowledge without accessing previous task data.

    % \item We show through extensive experiments on diverse segmentation domains that RegCL achieves favorable continual learning performance while preserving parameter efficiency and scalability.
    \item We demonstrate that RegCL delivers strong continual learning performance across diverse segmentation domains while preserving parameter efficiency and scalability.
\end{itemize}

\section{Related works}
\label{sec:related}

\subsection{Continual Learning}
Continual learning (CL), also known as lifelong learning, aims to learn a sequence of tasks while preserving knowledge acquired from previous ones~\cite{de2021continual}.
The central challenge is catastrophic forgetting, where adapting a model to new data causes severe degradation on earlier tasks.
Existing CL methods are commonly grouped into three families.

\emph{Regularization-based} methods preserve old knowledge by constraining important parameters or optimization directions during subsequent training.
Representative approaches estimate parameter importance through Fisher information, path-integral statistics, or output sensitivity, as in EWC~\cite{EWC}, Synaptic Intelligence~\cite{SI}, and MAS~\cite{MAS}.
Another line of work constrains gradient updates to reduce interference with previous tasks, including OGD~\cite{OGD}, GPM~\cite{GPM}, and CGP~\cite{CGP}.
Recent methods further combine gradient projection with adaptive merging mechanisms, e.g., BECAME~\cite{BECAME}, to improve the stability--plasticity trade-off.

\emph{Replay-based} methods mitigate forgetting by revisiting stored or generated historical samples.
Classical examples include iCaRL~\cite{icarl}, GEM~\cite{GEM}, and experience replay variants~\cite{ER}.
DER~\cite{DER} further improves rehearsal by matching logits from previous optimization trajectories, and has become a strong replay baseline in general continual learning.
Although effective, replay-based methods require storing or synthesizing historical data, which is undesirable when data privacy, storage, or domain access constraints are present.

\emph{Architecture-based} methods reduce task interference by allocating task-specific parameters or subnetworks.
Progressive Neural Networks~\cite{rusu2016progressive}, PackNet~\cite{PackNet}, and HAT~\cite{HAT} preserve previous knowledge by expanding, masking, or selectively reserving model capacity.
More recent parameter-efficient variants, such as O-LoRA~\cite{O_LoRA}, apply orthogonal low-rank subspaces to reduce interference in large-model continual learning.
However, these methods often require task-specific structures, masks, or adapters, and their storage or routing complexity may grow with the number of tasks.

Continual segmentation inherits these challenges but is more demanding, since the model must preserve dense pixel-level predictions.
Prior segmentation-oriented CL methods often employ distillation or class-incremental constraints to maintain previous segmentation behavior~\cite{michieli2019incremental}.
Nevertheless, these methods are not fully aligned with the adapter-based adaptation of foundation segmentation models.
Replay violates the non-replay assumption, strong regularization may restrict the plasticity of lightweight modules, and architecture expansion conflicts with the goal of maintaining a compact model across continuously emerging domains.
Different from these training-centric paradigms, RegCL studies a merging-based route: each new domain is adapted independently  and then consolidated into a compact running adapter without replay.

\subsection{Parameter-Efficient Fine-tuning}
Parameter-efficient fine-tuning (PEFT) adapts large pre-trained models by updating only a small number of parameters while keeping the backbone mostly frozen.
Representative methods include adapter modules~\cite{houlsby2019parameter}, LoRA~\cite{lora}, and AdaptFormer~\cite{adaptformer}.
They are well suited to vision transformers, where full fine-tuning is costly and storing one full model per domain is impractical.

PEFT has also become a common strategy for adapting SAM to specialized segmentation domains.
SAM-Adapter~\cite{SAM-Adapter} and Medical SAM Adapter~\cite{Sam-adapter-med} show that lightweight adaptation modules can substantially improve SAM on challenging datasets such as camouflaged, shadow, and medical image segmentation.
However, these methods are mainly single-domain solutions that produce one specialized adapter per target domain.

Continual SAM methods address this limitation with domain-specific module management.
CoSAM~\cite{moda} introduces a continual SAM benchmark and MoDA, a mixture of domain adapters, while SAMCL~\cite{samcl} uses domain-specific AugModules and a Module Selector for inference-time routing.
These methods are closely related to our setting because they address continual SAM adaptation.
However, they still rely on maintaining domain-specific modules or selection mechanisms, so the adaptation state grows as new domains accumulate.
In contrast, RegCL aims to continuously consolidate domain knowledge into a constant-size adapter through model merging, avoiding an expanding module library.

\subsection{Model Merging}
Model merging aims to combine multiple trained models into a single model without joint retraining.
Early studies show that weight-space interpolation can be effective when models share the same pre-trained initialization, as demonstrated by model soups~\cite{Soups}.
Task arithmetic~\cite{TA} further formulates fine-tuned model differences as task vectors, enabling model editing and capability composition through vector operations.
Beyond averaging, importance- or statistics-aware methods improve merging by accounting for parameter reliability or activation behavior.
For example, Fisher merging~\cite{finshermerging} weights parameters by Fisher information, while RegMean~\cite{Regmean} formulates model merging as a regression problem that preserves the predictions of constituent models.

A key difficulty in model merging is parameter interference.
Independently trained task updates may modify the same parameters in conflicting directions, making naive averaging unstable.
TIES~\cite{Ties} addresses this issue by trimming small updates, electing a dominant sign, and merges sign-consistent parameters.
DARE~\cite{dare} improves robustness by sparsifying and rescaling task vectors before merging.
These methods substantially improve multi-model fusion, but  typically require simultaneous access to all candidate models.

% Recently, model merging has been introduced into continual learning.
Recently, continual learning has begun to adopt model merging.
MagMax~\cite{magmax} leverages sequential fine-tuning and maximum-magnitude selection to accumulate knowledge across tasks.
DOP~\cite{DOP} studies data-free continual model merging and uses dual orthogonal projections to balance stability for previous tasks and plasticity for new tasks.
BECAME~\cite{BECAME} further connects adaptive merging with Bayesian continual learning.
% Despite these advances, most existing continual merging methods target full-model classification settings or data-free checkpoint fusion.
Most of them target full-model classification or data-free checkpoint fusion.
RegCL instead focuses on continual SAM adaptation with lightweight LoRA-style modules, maintaining compact RegMean-style feature statistics as historical state for non-replay dense segmentation.
\section{Method}
\label{sec:method}
%%%% background -> 改动
%%%% 1. 简单介绍SAM pipeline
%%%% 2. 我们的方法，定义一下SAM持续学习，形式化给出model merge持续学习算法（一堆公式）；只针对LoRA等训练参数做处理；画图，示意图（merge），pipeline图（持续学习）
%%%% 顺序无关

\begin{figure*}[ht]
    \centering
    \includegraphics[width=1\textwidth]{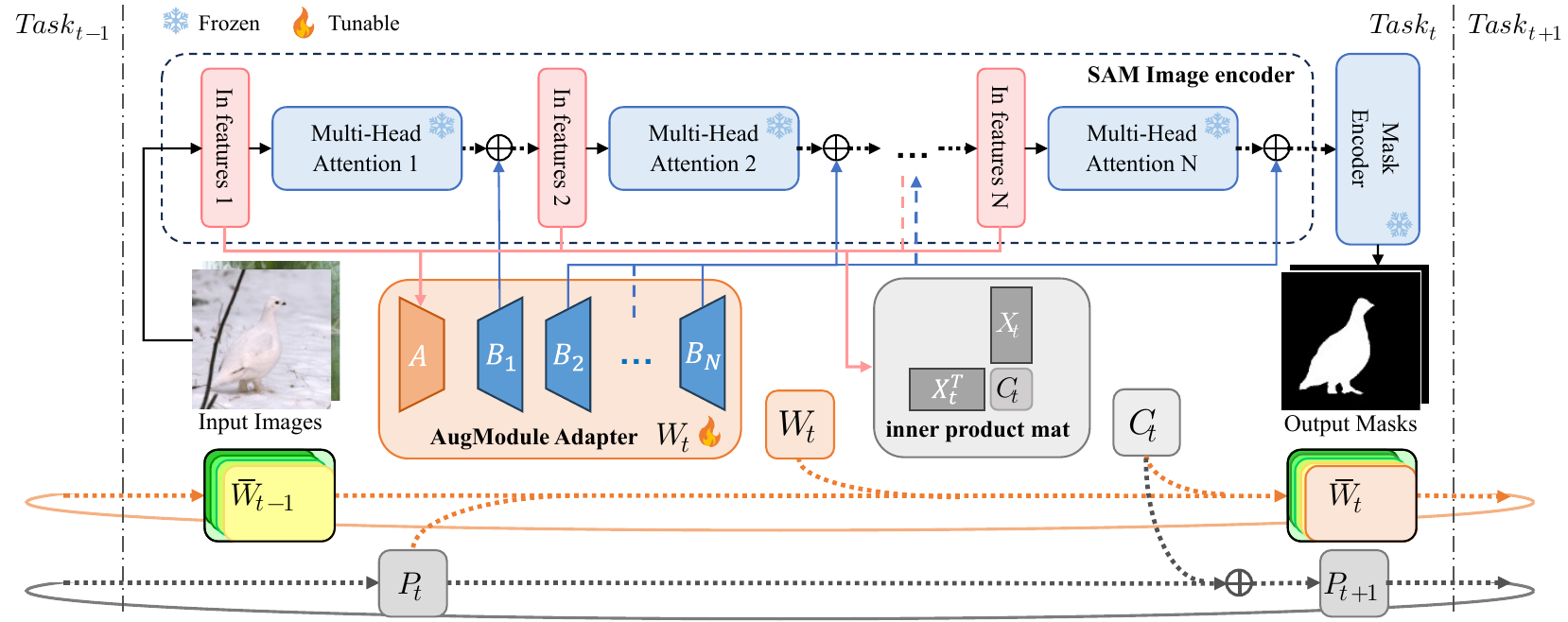}
    \caption{
    \textbf{The overall pipeline of RegCL.}
    After SAM is fine-tuned on a new task with LoRA modules.
    RegCL computes the feature inner product $C_i$  and updates an inner product accumulator $P_i$ used to merge current weights $W_t$ with previous weights $\overline{W}_{t-1}$. 
    The merged weights $\overline{W}_{t}$ are incrementally updated across tasks, enabling knowledge retention while adapting to new tasks.
    For details, please refer to Eq.~\eqref{eq:regcl}--~\eqref{eq:regcl_avg}.
    }
    \label{fig:pipeline}
\end{figure*}

%一段话：问题陈述+方法整体架构描述
We present \textbf{Regression Continual Learning (RegCL)}, a novel non-replay CL method by leveraging adaptive parameter merging to balance historical and new task knowledge. The approach integrates task-specific training with a dynamic parameter merging mechanism formulated as a linear regression problem, ensuring effective knowledge retention while adapting to new tasks.

\subsection{Problem Setup}
We consider a problem of continual learning of LoRA adapter modules.
Specifically, our task setup can be categorized as Domain-Incremental Learning (DIL), where tasks share the same data label space but have different input distributions, and task identities are not provided~\cite{DILa}~\cite{DILb}.
In DIL, we aim to adapt a single LoRA adapter module to a sequence of tasks $\{\mathcal{D}_1, \mathcal{D}_2, \dots, \mathcal{D}_T\}$ sequentially. 
Notably, we follow a non-replay protocol that prohibits access to data from previous tasks, preventing storage requirements from increasing as the number of tasks grows.

\subsection{Continual Parameter Merging}
%一段话：

% \noindent \paragraph\textbf{{}}

% \noindent \paragraph\textbf{{Bounded Historical State.}}

% \noindent \paragraph\textbf{{}}

% We formulate parameter merging as an optimization problem that minimizes the output discrepancy between models with the same architecture.
% 
We derive the continual update by preserving the functional responses of task-specific adapters. For a linear transformation, parameters that produce similar outputs on the corresponding task features can be regarded as functionally compatible even when their element-wise values differ. Accordingly, we seek a shared parameter matrix whose responses remain close to those of the task-specific transformations on their respective feature distributions.

Consider two linear models $f_1(x) = W_1^\top x$ and $f_2(x)= W_2^\top x$, where $x \in \mathbb{R}^m$, and $W_1, W_2 \in \mathbb{R}^{m\times n}$, that are trained on two different datasets, $\langle X_1, y_1 \rangle$ and $\langle X_2, y_2 \rangle$, where $X_1 \in\mathbb{R}^{N_1 \times m}$ and $X_2 \in \mathbb{R}^{N_2 \times m}$ are input.
Each row of $X_{i}$ corresponds to a training example. The objective is to obtain a single merged model $f_M(x)=W_{M}^{T} x$, whose outputs approximate $f_1$ on $X_1$ and $f_2$ on $X_2$.
Using $\ell^2$-distance metric, the optimization problem is expressed as:
\begin{equation}
    \label{eq:opt_problem}
    \min_{W} \quad \lVert W^\top X_1 - W_1^\top X_1 \rVert^2 + \lVert W^\top X_2 - W_2^\top X_2 \rVert ^2.
\end{equation}
This formulation represents a linear regression problem where the inputs are $[X_1; X_2]$ (row-wise concat of $X_1$ and $X_2$), and the targets are $[W_1^\top X_1; W_2^\top X_2]$. The closed-form solution to this optimization problem is:
% \begin{equation}
\begin{align}
\label{eq:merge_2_linear}
\nonumber
    W_{M_{1,2}} &= (X_1^\top X_1 + X_2^\top X_2)^{-1} (X_1^\top X_1W_1 + X_2^\top X_2W_2)\\
    &=(C_1+C_2)^{-1}(C_1W_1 + C_2W_2),
\end{align}
where $C_i=X_i^\top X_i$.

Applying the same response-preservation objective to a collection of task-specific transformations indexed by $\mathcal{K}$ yields
\begin{equation}
    \label{eq:merge_multi_linear}
    % W_M = (\sum_i^{i \in \mathcal{K}} X_i^\top X_i)^{-1} \sum_i^{i \in \mathcal{K}}(X_i^\top  X_i W_i).
    W_M = (\sum_i^{i \in \mathcal{K}} C_i)^{-1} \sum_i^{i \in \mathcal{K}}(C_i W_i),
\end{equation}
showing that each task contributes through the pair $(C_i,W_i)$.
The task data themselves are therefore unnecessary once the feature statistics $C_i$ have been accumulated.

We further adapt this merging formulation to continual learning, where task-specific models from previous tasks are no longer simultaneously available.
When the current task $\mathcal{D}_t$ arrives, we retain only the previously merged parameters $\overline{W}_{t-1}$ and an accumulated inner-product matrix summarizing the preceding tasks.
Let $\overline{W}_t$ denote the merged parameters after learning the first $t$ tasks.
Applying Eq.~\ref{eq:merge_multi_linear} to these tasks and separating the current task from the preceding tasks gives
\begin{equation}
    \label{eq:merge_multi_linear_t}
    \overline{W}_t
    =
    \left(
        \sum_{i=1}^{t-1}C_i+C_t
    \right)^{-1}
    \left(
        \sum_{i=1}^{t-1}C_iW_i+C_tW_t
    \right).
\end{equation}

We define historical inner-product accumulator as $P_t=\sum_{i=1}^{t-1}C_i$.
According to the definition of $\overline{W}_{t-1}$ in Eq.~\ref{eq:merge_multi_linear}, $P_t\overline{W}_{t-1}=\sum_{i=1}^{t-1}C_iW_i$.
Therefore, the contributions of all previous task-specific parameters can be represented exactly by the pair $(P_t,\overline{W}_{t-1})$, without retaining the individual historical models.
Substituting this relation into Eq.~\ref{eq:merge_multi_linear_t} yields the continual merging rule of RegCL:
\begin{equation}
    \overline{W}_t
    = \underbrace{\left(P_t+C_t\right)^{-1}}_{\text{Adaptive weighting}}
    \left(
        \underbrace{P_t\overline{W}_{t-1}}_{\text{Historical knowledge}}
        +
        \underbrace{C_tW_t}_{\text{New knowledge}}
    \right).
    \label{eq:regcl}
\end{equation}

The three components in Eq.~\ref{eq:regcl} have distinct roles.
First, $P_t\overline{W}_{t-1}$ represents historical knowledge because
$P_t$ aggregates the input statistics of all previous tasks, while
$\overline{W}_{t-1}$ encodes their merged parameters.
More precisely, their product is equivalent to the complete historical contribution
$\sum_{i=1}^{t-1}C_iW_i$.
Importantly, this storage cost remains constant with respect to the number of observed tasks because we maintain only the cumulative matrices rather than one matrix per task.
The exact overhead relative to LoRA parameters depends on the adapter rank and the dimensions of the selected transformations.

Second, $C_tW_t$ represents new knowledge introduced by the current task.
Here, $W_t$ contains the parameters learned from $\mathcal{D}_t$, whereas $C_t$ describes the input-feature statistics under which these parameters are optimized.
Consequently, the current parameters contribute more strongly along feature directions that are more relevant to the current task.

Furthermore, the feature-statistics matrices $C_t$ can be accumulated through one additional forward pass after fine-tuning.
In contrast, estimating gradient-based importance measures such as Fisher information generally requires backward computation~\cite{finshermerging}.
By doing so we avoid additional gradient computation associated with such importance estimators.

After computing the feature-statistics matrix $C_t$, the raw samples and annotations of task $\mathcal{D}_t$ can be discarded, as subsequent merging only requires $C_t$. This avoids storing or revisiting task-specific data, thereby improving data privacy.

Third, $(P_t+C_t)^{-1}$ provides adaptive weighting by normalizing the historical and current contributions according to their aggregated input statistics.
Unlike scalar interpolation coefficients, matrix-valued weighting operates differently across feature directions.
Directions strongly supported by historical tasks are primarily constrained by $P_t$, whereas directions that are informative for the current task receive a larger contribution through $C_t$.
% This interpretation can also be seen by rewriting Eq.~\ref{eq:regcl} as
% \begin{equation}
%     \overline{W}_t
%     =
%     \overline{W}_{t-1}
%     +
%     \left(P_t+C_t\right)^{-1}C_t
%     \left(W_t-\overline{W}_{t-1}\right).
%     \label{eq:regcl_update}
% \end{equation}
% The matrix $(P_t+C_t)^{-1}C_t$ therefore acts as a task-dependent update coefficient that determines how strongly the merged parameters move from the historical solution toward the current task-specific solution.

After obtaining $\overline{W}_t$, the historical statistics are updated for the next task as:
\begin{equation}
    P_{t+1}
    =
    \sum_{i=1}^{t}C_i
    =
    P_t+C_t.
    \label{eq:regcl_p}
\end{equation}

This recursive update retains the cumulative input statistics of all observed tasks in $P_{t+1}$ without storing raw data or task-specific inner-product matrices. Thus, $\overline{W}t$ and $P{t+1}$ suffice to process the next task $\mathcal{D}_{t+1}$.

For parameters in nonlinear components, for which the linear regression formulation in Eq.~\ref{eq:opt_problem} does not directly apply, we use an incremental arithmetic average:
\begin{equation}
    \overline{W}_t
    =
    \frac{1}{t}
    \left(
        (t-1)\overline{W}_{t-1}+W_t
    \right).
    \label{eq:regcl_avg}
\end{equation}
This update is equivalent to uniformly averaging the corresponding parameters learned from all tasks observed up to time step $t$ and requires only the previously averaged parameters.

\subsection{Regression Continual Learning}
% pipeline explian start
The overall RegCL pipeline is illustrated in Figure~\ref{fig:pipeline}.
We freeze the pretrained SAM backbone and introduce trainable LoRA adapters into its image encoder.
Specifically, we adopt the AugModule adapter from SAMCL~\cite{samcl}.
Each task-specific adapter $W_t$ contains $N$ matrices $\{B_i\}_{i=1}^{N}$ and one parameter-shared matrix $A$.
The shared matrix $A$ is paired with each $B_i$ to construct $N$ LoRA modules.

At time step $t$, we obtain a task-specific adapter $W_t$ on the current dataset $\mathcal{D}_t$ using either independent or sequential fine-tuning.
After fine-tuning, we perform an additional forward pass over $\mathcal{D}_t$ and collect the input features $X_t$ of each mergeable linear transformation.
The corresponding feature-statistics matrix is then computed as $C_t=X_t^\top X_t$.
Once $C_t$ has been accumulated, the raw samples and intermediate activations of $\mathcal{D}_t$ can be discarded.
RegCL therefore retains only the task-specific parameters and compact feature stati0stics required by the subsequent merging step.

For the first task, no historical knowledge is available.
We therefore initialize the continual states as:
\begin{equation}
    \overline{W}_1=W_1,
    \qquad
    P_2=C_1.
\end{equation}
For every subsequent task $t>1$, the newly trained parameters $W_t$ are merged with the historical parameters $\overline{W}_{t-1}$ using Eq.~\ref{eq:regcl}.
The accumulator is then updated using Eq.~\ref{eq:regcl_p}, and the resulting states $(\overline{W}_t,P_{t+1})$ are retained for the next task.

This procedure is repeated over the complete task sequence
$\{\mathcal{D}_1,\ldots,\mathcal{D}_T\}$.
After the final task, RegCL returns
\begin{equation}
    W_M=\overline{W}_T.
\end{equation}
The resulting adapter integrates the knowledge acquired from all observed tasks into a single parameter set.

The complete RegCL procedure is summarized in Algorithm~\ref{alg:regcl}.
\begin{table*}[t]
\centering

\caption{
\textbf{Domain-incremental learning performance across five datasets (Kvasir $\rightarrow$ CAMO $\rightarrow$ ISTD $\rightarrow$ ISIC $\rightarrow$ COD).} 
All methods use the same backbone and fine-tuning protocol.
% 
% 'LoRA-Seq' denotes sequential fine-tuning with LoRA adapters.
% 
% 'Upper bound',which is only here for reference, refers to the average results obtained by fine-tuning and testing the adapter on each dataset separately, which corresponds to a 0\% performance drop across domains. '-' indicates that non-CL methods cannot produce cross-task metrics (e.g., BWT and FWT).
% 
Best results are in \textbf{bold}, and best order-independent results are \underline{underlined}.
RegCL achieves the best overall performance, demonstrating competitive forward transfer ability compared to other methods.
}\label{tab:main_res}

% \resizebox{\textwidth}{!}{
\begin{tabular}{c||ccc|ccc|ccc}
\hline
\multicolumn{10}{c}{Kvasir $\rightarrow$ CAMO $\rightarrow$ ISTD $\rightarrow$ ISIC $\rightarrow$ COD} \\ \hline
\multirow{2}{*}{Method} & \multicolumn{3}{c|}{ACC}& \multicolumn{3}{c|}{BWT} & \multicolumn{3}{c}{FWT} \\ 
\cline{2-10}
& mIoU $\uparrow$ & mF1$\uparrow$ & mMAE$\downarrow$
& mIoU$\uparrow$ & mF1$\uparrow$ & mMAE $\downarrow$
& mIoU$\uparrow$ & mF1$\uparrow$ & mMAE $\downarrow$\\ \hline 

% \textbf{Lora~\cite{lora}}
% & 0.703& 0.804&	0.064
% & 0.123& 0.089&	-0.034
% & 0.513& 0.521& 0.158 \\
SAM~\cite{SAM}
& 0.647& 0.755&	0.098
& -& -&	-
& -& -& - \\

SAM3~\cite{SAM3}
& 0.745& 0.827&	0.077
& -& -&	-
& -& -& - \\

% LoRA-Mean
% & 0.739& 0.831&	0.055
% & -& -&	-
% & -& -& - \\

LoRA-Seq~\cite{lora}
& 0.691& 0.798&	0.061
& -0.125& -0.089& 0.029
& 0.528& 0.647& 0.176 \\

EWC~\cite{EWC}
& 0.716& 0.816&	0.058
& -0.111& -0.078&	0.028
& 0.549& 0.663& 0.160 \\
SPPA~\cite{SPPA}
& 0.282& 0.407& 0.149
& -0.337& -0.315& 0.072
& 0.417& 0.550& 0.197\\
LAG~\cite{LAG}
& 0.703& 0.810&	0.063
& -0.099& -0.066& 0.025
& 0.452& 0.576& 0.205 \\
O-LoRA~\cite{O_LoRA}
& 0.704& 0.806& 0.059
& -0.091& -0.066& 0.023
& 0.519& 0.642& 0.160\\
\hline
% ER~\cite{ER}
% & 0.808& 0.881&	\underline{0.035}
% & \underline{-0.010}& \underline{-0.007}& \underline{0.003}
% & 0.630& 0.748&	0.087\\
% DER~\cite{DER}
% & 0.804& 0.879&	\underline{0.035}
% & -0.022& -0.015&	0.005
% & 0.643& 0.760&	\underline{0.082} \\
TIES~\cite{Ties}
& 0.736& 0.828&	0.058
& \textbf{0.000}& \textbf{0.000}& \textbf{0.004}
& 0.665& 0.774&	0.073\\
Magmax~\cite{magmax}
& 0.741& 0.832&	0.056
& -0.025& -0.019& 0.010
& \textbf{0.678}& \textbf{0.784}& \textbf{0.069}\\
DOP~\cite{DOP}
& 0.771& 0.855&	0.047
& -0.038& -0.026& 0.012
& 0.623& 0.735&	0.096\\
\hline

\textbf{RegCL}& 
% \textbf{0.751}& \textbf{0.840}& \textbf{0.048}& \textbf{-0.028}& \textbf{-0.021}& \textbf{0.006}& \textbf{0.651}& \textbf{0.763}& \textbf{0.084} \\
\underline{0.759}& \underline{0.846}& \underline{0.048}& \underline{-0.024}& \underline{-0.018}& \underline{0.008}& \underline{0.659}& \underline{0.771}& \underline{0.074} \\
\textbf{RegCL+Seq}
& \textbf{0.775}& \textbf{0.856}&	\textbf{0.046}
& -0.007& -0.005& 0.005
& 0.668& 0.777&	0.071\\
% \textbf{RegCL+Replay}
% & \underline{0.809}& \underline{0.882}& 0.036
% & -0.018& -0.013& 0.005
% & \underline{0.651}& \underline{0.764}& 0.084 \\
\textbf{Upper Bound}&
0.820& 0.890& 0.030&
-& -& -&
-& -& -\\
\hline

\end{tabular}
% \caption{\textbf{Domain-incremental learning performance comparison across five datasets (Kvasir $\rightarrow$ CAMO $\rightarrow$ ISTD $\rightarrow$ ISIC $\rightarrow$ COD)}. 
% % 
% `LoRA-Seq' denotes the sequential learning with LoRA adapters.
% % 
% “Upper bound” refers to the average results obtained by fine-tuning and testing the adapter on each dataset separately, which corresponds to a 0\% performance drop across domains.
% % 
% All methods share the same fine-tuning architecture and training strategy.
% % 
% The best results are shown in \textbf{bold} for non-replay based methods, and in \underline{underline} for replay based methods.
% % 
% Overall, RegCL achieves the best performance.
% }
\end{table*}
\begin{algorithm}
    \caption{Pseudo Code of RegCL}
    \label{alg:regcl}
    \SetAlgoLined
    \DontPrintSemicolon
    \KwIn{
        Task $\mathcal{D}=\{\mathcal{D}_1, \mathcal{D}_2, \dots, \mathcal{D}_T\}$\;
        Initial parameters $W_0$ (kaiming initialization)\;
    }
    \KwOut{Merged parameters $W_M$}
    
    \textbf{Initialize:}\;
    \Indp
        Inner product accumulator $P \leftarrow \mathbf{0}$\;
        % Parameter memory $W_{\text{prev}} \leftarrow W_0$\;
    \Indm
    
    \For{$\mathcal{D}_t \in \{\mathcal{D}_1, \mathcal{D}_2, \dots, \mathcal{D}_T\}$}{
        \textbf{Step a: Train Task-Specific Model}\;
        
        \Indp
        Initialize $W_t \leftarrow W_0$
        
        \For{$\{x,y\} \in D_t$}{
            Update $W_t$ with $\mathcal{L}_{Total}$ in Eq.~\ref{eq:Loss_s}
        }
         Compute inner product matrix $\mathbf{C}_t = X_t^\top X_t$\;
        \Indm
        
        \textbf{Step b: Merge Parameters via RegCL}\;
        \Indp
        % \If{$t > 1$}{
        %     Merge parameters:
        %     $\overline{W}_{t}=(P_t+C_t)^{-1}(P_t\overline{W}_{t-1}+C_tW_t)$.
        %     % W_{\text{agg}} = \underbrace{(P+C_t)^{-1}}_{\text{Adaptive weighting}} \left( \underbrace{P W_{\text{prev}}}_{\text{Historical knowledge}} + \underbrace{C_t W_t}_{\text{New knowledge}} \right)
            
        %     For non linear layer weights in $\overline{W}_{t-1}$ and $W_t$, average weights as:
        %     $\overline{W}_{t}=\frac{1}{t}((t-1)\times\overline{W}_{t-1}+W_t).$
        % }
        
        % $\overline{W}_{t} \leftarrow W_t$
        % (First task need no merging)
        
        \eIf{$t == 1$}{
            $\overline{W}_{t} \leftarrow W_t$ (First task need no merging)
        }
        {
            Merge parameters:
            $\overline{W}_{t}=(P_t+C_t)^{-1}(P_t\overline{W}_{t-1}+C_tW_t)$.
            % W_{\text{agg}} = \underbrace{(P+C_t)^{-1}}_{\text{Adaptive weighting}} \left( \underbrace{P W_{\text{prev}}}_{\text{Historical knowledge}} + \underbrace{C_t W_t}_{\text{New knowledge}} \right)
            
            For non linear layer weights in $\overline{W}_{t-1}$ and $W_t$, average weights as:
            $\overline{W}_{t}=\frac{1}{t}((t-1)\times\overline{W}_{t-1}+W_t).$ 
        }

        $P_{t+1} = P_t + \mathbf{C}_t$ (Inner product matrix prepared for next merging)\;
        \Indm
    }
    % \Return{$W_M=W_T$}
\end{algorithm}
% pipeline explian ends

\subsubsection{Sequential Fine-tuning}
RegCL supports both independent and sequential fine-tuning.
With independent fine-tuning, each task-specific adapter $W_t$ is trained from the same initialization using only $\mathcal{D}_t$.
As shown in Equ~\ref{eq:merge_multi_linear}, the final merged parameters depend only on the sums $\sum_{t=1}^{T} C_t$ and $\sum_{t=1}^{T} C_tW_t$, which are invariant to task permutations, this configuration is independent of the task order. Its robustness across different task sequences is evaluated in Table~\ref{tab:order_robustness}. A more comprehensive analysis is provided in ~\ref{subsubsec:order-robustness}. 

Independently fine-tuned models can reach functionally similar solutions through inconsistent parameter-update directions, as over-parameterized networks often admit multiple equivalent optima.
The resulting sign conflicts may cause task-specific updates to cancel or interfere during merging~\cite{Ties}.
Sequential fine-tuning has been empirically shown to reduce such conflicts by keeping successive models within a more compatible parameter region~\cite{magmax}.
We therefore use sequential fine-tuning as the default RegCL configuration, while retaining independent fine-tuning as an alternative when order-invariant task training is preferred.

%%% samcl copy
\subsubsection{Fine-tuning Loss}
Following SAM \cite{SAM}, during fine-tuning of task-specific SAM adapters, we employ MSE loss, focal loss~\cite{focalloss}, and dice loss~\cite{diceloss} to supervise the mask prediction. We set hyperparameter to 10 following previous SAM finetuning methods.
The overall loss function is formulated as:
\begin{equation}
    \mathcal{L}_{Total} = \mathcal{L}_{MSE} + \mathcal{L}_{Focal} + 10 \times \mathcal{L}_{Dice}.
  \label{eq:Loss_s}
\end{equation}

% During fine-tuning of task-specific SAM adapters, we follow the original implementation in SAM \cite{SAM}, modifying only the coefficients. 
% %
% The IoU prediction head is trained with mean squared error loss, which measures the difference between the predicted IoU and the IoU of the predicted mask compared to the ground truth mask. 
% %
% We supervise the mask prediction using a linear combination of focal loss \cite{focalloss} and dice loss \cite{diceloss}. 
% The complete loss function is formulated as:
% \begin{equation}
%     \mathcal{L}_{Total} = \mathcal{L}_{Focal} + 10 * \mathcal{L}_{Dice} + \mathcal{L}_{Mse}.
%   \label{eq:Loss_s}
% \end{equation}
%%% samcl copy
% \input{table/order}

\section{Experiment}
\label{sec:expriment}

% Move the replay table float earlier while preserving its printed table number.
% \addtocounter{table}{4}
% \input{table/ablation_replay}
% \addtocounter{table}{-5}
\subsection{Implementation Details}

\noindent\textbf{Datasets.}
% We evaluate the effectiveness of RegCL on five datasets across three common applications domains of SAM adapters: medical images, shadows and camouflaged objects.
We evaluate the effectiveness of RegCL on five visually heterogeneous segmentation datasets that stress the visual analysis layer of multi-sensorial media systems, including medical imagery, shadow-dominant scenes, and camouflaged objects.
For medical images, we use \textbf{Kvasir}~\cite{kvasir} and \textbf{ISIC}~\cite{ISIC} for gastrointestinal polyp and skin lesions images.
For camouflaged objects, we adopt \textbf{CAMO}~\cite{camo} and \textbf{COD}~\cite{COD}, which cover diverse scenarios, including land, sea, and air creatures, as well as challenges like background clutter, object occlusion and etc.
For shadow images, we adopt \textbf{ISTD}~\cite{ISTD}, which provides paired shadow image and mask.

\noindent\textbf{Fine-tuning.}
To demonstrate RegCL’s effectiveness, we aim to preserve SAM’s generalization ability while adapting it to diverse segmentation tasks.
We use SAM with ViT-B/16 backbone as the segmentation model.
During fine-tuning, only the weights of the adapter module attached to the image encoder is updated, while all other weights remain frozen.
% 
% To reduce computation and efficiently extract dataset features for the inner product $C_t$ s, we consolidate the low-rank $A$ of each layer into a single entity. 
Specifically, we adopt AugModule from SAMCL~\cite{samcl}, as research has shown that it can outperform LoRA while using fewer parameters.
The mask decoder and prompt encoder are kept frozen and directly integrated into our framework, with point-type prompts used for the prompt encoder.
% sam 在实验当中中的具体实现方法：
% 在这项工作中，我们的目标是利用并保留 “任意分割模型”（SAM）的泛化能力，同时使其有效地适应各种分割任务。为此，我们将 SAM 作为分割框架的支柱。
% 我们选择在 SAM 中使用的图像编码器是 ViT-B/16 模型，该模型将图像划分为 16×16 补丁，采用 12 层 Transformer 架构，带有多头自注意和前馈网络。每一层都能捕捉图像中的全局依赖关系，从而生成高质量的特征表示。
%在适配过程中，我们冻结了预训练图像编码器和原始解码器的权重，仅在图像编码器上采用 LoRA 模块。
% 
% 至于掩码解码器和提示解码器，我们直接将其纳入我们的框架，不做任何修改。此外，我们还在原有的提示解码器中采用了点类型提示。
% 
We fine-tune SAM for 20 epochs with a batch size of 8 per dataset. The initial learning rate is 0.005 with the Cosine Annealing schedule.

\noindent\textbf{Metrics.}
Segmentation quality is measured by three common metrics, mean absolute error (mMAE), F1 score (mF1), and intersection over union (mIoU).
To evaluate the performance on continual learning, we follow GEM~\cite{GEM} to adopt three metrics as follows: 1) Average Accuracy (ACC) is defined as ACC $=\frac{1}{T}\sum_{i=1}^T R_{T,i}$; 2) Backward Transfer (BWT) is defined as BWT $=\frac{1}{T}\sum_{i=1}^{T-1} R_{T,i}-R_{i,i}$; 3) Forward Transfer (FWT) is defined as FWT $=\frac{1}{T-1}\sum_{i=1}^{T-1} R_{i, i+1}$, where $R_{i,j}$ represents the accuracy for the \textit{j}-th task after training on the \textit{i}-th task.
% $a_{i,j}$ represents the accuracy for the \textit{j}-th task after training on the \textit{i}-th task.
% To evaluate the performance in terms of learning, maintaining, and transferring knowledge, we follow the approach of previous works 
% % \cite{GEM, 2023survey}
% and define three key metrics as follows: 1) Average Accuracy (ACC) is defined as ACC $=\frac{1}{T}\sum_{i=1}^T a_{T,i}$; 2) Forgetting Measure (FM) is defined as FM $=\frac{1}{T}\sum_{i=1}^{T} a_{i,i} - a_{T,i}$; 3) Forward Transfer (FT) is defined as FT $=\frac{1}{T-1}\sum_{i=1}^{T-1} a_{i, i+1}$.
% $a_{i,j}$ represents the accuracy for the \textit{j}-th task after training on the \textit{i}-th task. To evaluate segmentation performance, we employ three common metrics: mean absolute error (mMAE), mean F1 score (mF1), and mean intersection over union (mIoU).

\subsection{Main Results}
We evaluate the proposed method on the five datasets in a continual learning setting. The order of the datasets is Kvasir, CAMO, ISTD, ISIC, then COD.
In Table~\ref{tab:main_res}, bold values indicate the best non-replay results, and underlined values indicate the best order-independent results.
We compare RegCL with foundation-model baselines, the simple sequential LoRA baseline, conventional non-replay continual learning methods, and model-merging continual learning methods.
All methods were evaluated under the same backbone and fine-tuning protocol to ensure a fair comparison.

LoRA-Seq is adopted as the most direct sequential adaptation baseline, where the LoRA adapter is fine-tuned along the task stream without an explicit retention mechanism.
Compared with LoRA-Seq, RegCL improves Average Accuracy by 0.068 mIoU and 0.048 mF1, while reducing mMAE by 0.013.
This shows that simply continuing to update the adapter is insufficient for preserving previously acquired domain knowledge.

The Upper Bound row is reported only as a reference: it averages the results obtained by independently fine-tuning and testing an adapter on each dataset, corresponding to an ideal setting with no cross-domain performance drop and no continual transfer metrics.
RegCL narrows the gap to this reference while maintaining a single consolidated adapter across the task stream.

\noindent\textbf{Conventional CL.}
We compare with representative non-replay CL methods, including EWC~\cite{EWC}, SPPA~\cite{SPPA}, LAG~\cite{LAG}, and O-LoRA~\cite{O_LoRA}.
Compared with the best conventional CL baseline, RegCL improves ACC by 0.043 mIoU and 0.030 mF1, while reducing mMAE by 0.010.
For Backward Transfer, RegCL achieves -0.024 mIoU, -0.018 mF1, and 0.008 mMAE, indicating that the model only suffers minor performance degradation after learning all domains.

\noindent\textbf{Merging CL.}
TIES~\cite{Ties}, MagMax~\cite{magmax}, and DOP~\cite{DOP} are included as representative model-merging baselines to assess the effectiveness of alternative fusion strategies in comparison with RegCL.
MagMax achieves strong Forward Transfer, suggesting that merged weights can benefit zero-shot adaptation to subsequent tasks. And Ties-merging achieves strong Backward Transfer, since their method is specifically designed to address sign conflicts.
Nevertheless, ACC remains our primary metric for comparison.
RegCL provides the best order-independent results, and RegCL+Seq achieves the strongest ACC among all compared non-replay methods while remaining competitive in both BWT and FWT.

\noindent\textbf{Pre-trained Models.}
Beyond CL baselines, we further compare with SAM3~\cite{SAM3} as a recent foundation segmentation model.
Though SAM3 improves over the original SAM, RegCL with SAM1 still achieves higher ACC across all three segmentation metrics, indicating that continual adapter consolidation can be more effective than directly relying on a stronger generic foundation model in this specialized domain-incremental setting.

Results demonstrate that RegCL effectively balances knowledge retention, new-domain adaptation, and transfer to unseen domains in the CL setting.
This balance is important when the visual grounding component of a media system must be updated under changing deployment conditions.

\begin{figure}[ht]
    \includegraphics[width=\linewidth]{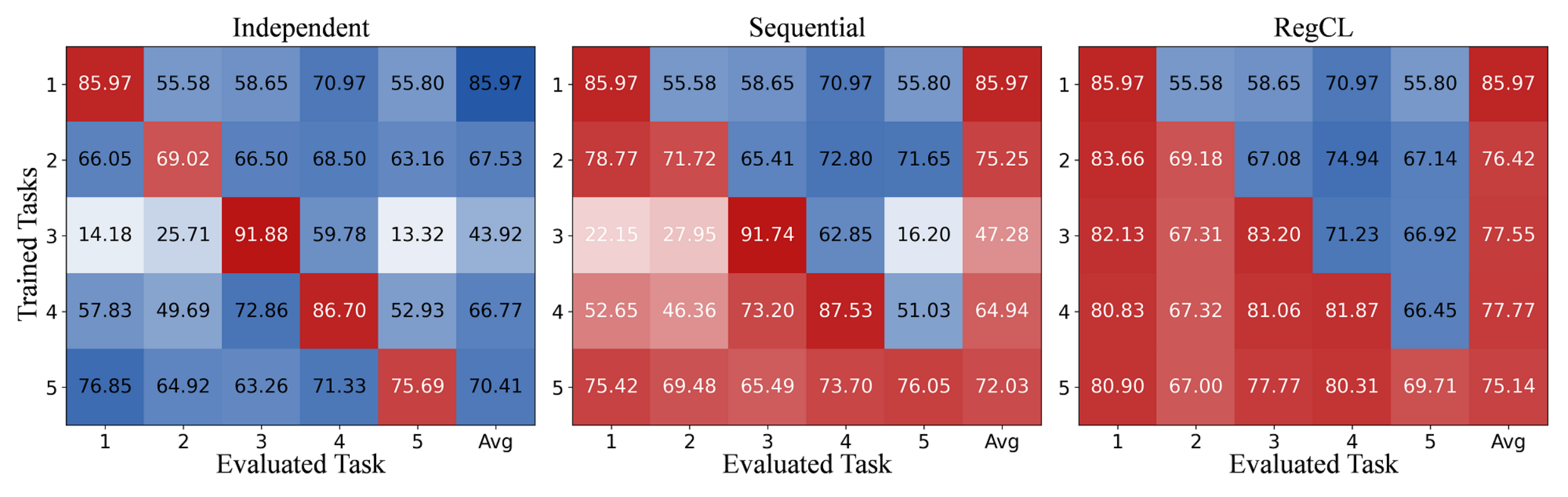}
    \caption{
    \textbf{Continual segmentation performance across tasks.}
    Red entries denote learned tasks, and blue entries denote zero-shot evaluation on unseen tasks.
    }
    \label{fig:ism_comp}
\end{figure}

% Furthermore, Figure~\ref{fig:ism_comp} presents the accuracy of independent fine-tuning, LoRA sequential fine-tuning, and our RegCL for each dataset during the continual learning process.  
Furthermore, Figure~\ref{fig:ism_comp} presents the accuracy of independent fine-tuning, LoRA sequential fine-tuning, and RegCL during the continual learning process.
% 
% Sequence fine-tuning exhibits catastrophic forgetting of previously learned tasks, aligning with previous studies.
In each matrix, columns denote evaluation datasets and rows denote the model state after each training step.
The final column reports the average accuracy over seen tasks, computed from the lower triangular region.
Sequential fine-tuning exhibits catastrophic forgetting of previously learned tasks, aligning with previous studies.
% 
% In contrast, RegCL preserves significantly more performance on earlier tasks after learning new ones, demonstrating stronger task retention and more balanced performance across domains. 
In contrast, RegCL preserves more performance on earlier tasks after learning new domains, demonstrating stronger retention and more balanced performance across domains.

\begin{figure}
    \includegraphics[width=1.0\linewidth]{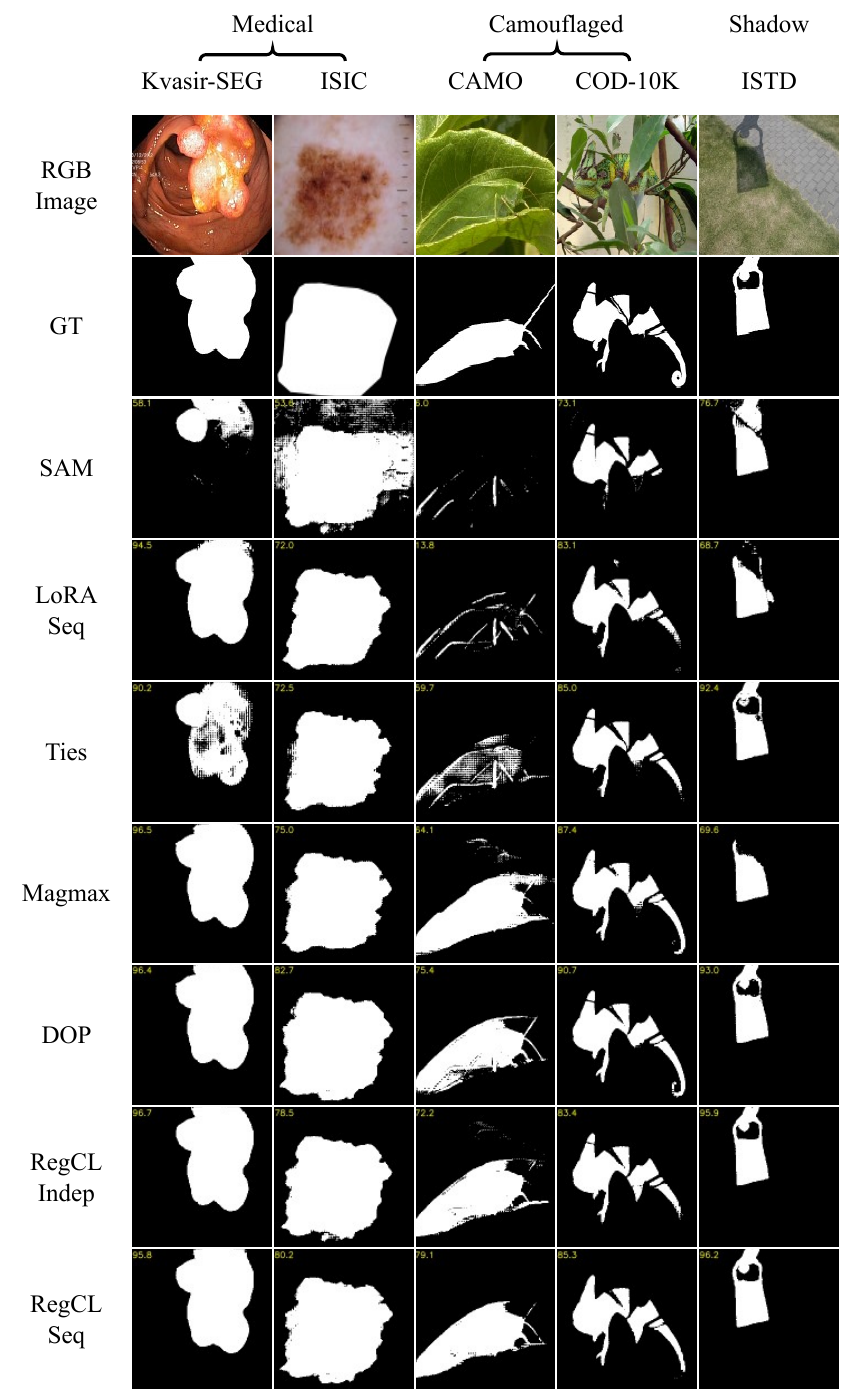}
    \caption{
    \textbf{Visualization results of the segmentation mask.} From left to right, we show results on medical segmentation (Kvasir-SEG, ISIC), camouflaged objects segmentation (CAMO, COD-10K), and shadow object segmentation (ISTD).
    }
    \label{fig:vis_comp}
\end{figure}

\subsection{Visualization Analysis}
For qualitative evaluation, Figure~\ref{fig:vis_comp} presents segmentation results from SAM, RegCL, and other methods across all categories, alongside the input RGB images and ground-truth (GT) masks.
As observed , SAM struggles in challenging scenarios, while RegCL produces more accurate and complete masks across all datasets.
This improvement comes from better capturing domain-specific features (e.g., mucosal textures, lesion boundaries, and shadow continuity), leading to smoother and more coherent segmentation.

\subsection{Ablation}
% \begin{table*}[ht]
\begin{table*}[!ht]
\centering
\caption{
\textbf{Domain-incremental learning performance with replay samples.}
% The performance of RegCL can be further improved with replay samples. 
% Comparing result in CS setting with an interspersed order. All methods are executed on the SAM within 20 epochs for each task.
% \ourM~outperforms all other methods. After finishing CS, the final testing results for all datasets are provided in the Appendix \ref{sec:different_order}.
}
\scalebox{0.9}{
\begin{tabular}{c||ccc|ccc|ccc}
\hline
\multicolumn{10}{c}{Kvasir $\rightarrow$ CAMO $\rightarrow$ ISTD $\rightarrow$ ISIC $\rightarrow$ COD} \\ \hline
\multirow{2}{*}{Method}
& \multicolumn{3}{c|}{ACC}
& \multicolumn{3}{c|}{BWT}
& \multicolumn{3}{c}{FWT} \\ 
\cline{2-10}
& mIoU $\uparrow$ & mF1$\uparrow$ & mMAE$\downarrow$
& mIoU$\uparrow$ & mF1$\uparrow$ & mMAE $\downarrow$
& mIoU$\uparrow$ & mF1$\uparrow$ & mMAE $\downarrow$\\ \hline 

\textbf{ER~\cite{ER}}
& 0.808& 0.881&	\textbf{0.035}
& \textbf{-0.010}& \textbf{-0.007}& \textbf{0.003}
& 0.630& 0.748&	0.087\\
\textbf{DER~\cite{DER}}
& 0.804& 0.879&	\textbf{0.035}
& -0.022& -0.015&	0.005
& 0.643& 0.760&	0.082 \\

\hline

\hline
\textbf{RegCL}& 
0.759& 0.846& 0.048&
-0.024& -0.018& 0.008&
\textbf{0.659}& \textbf{0.771}& \textbf{0.074} \\
% \textbf{RegCL+Repaly}
\textbf{RegCL+Replay}
& \textbf{0.809}& \textbf{0.882}& 0.036
& -0.018& -0.013& 0.005
& 0.651& 0.764& 0.084 \\
% \textbf{RegCL+repalyB (Ours)}
% & 0.747& 0.835& 0.050
% & 0.061& 0.045& -0.015
% & 0.621& 0.726& 0.118 \\
\hline

\end{tabular}}
\label{tab:ab_replay}
\end{table*}

\begin{table}[ht]
\caption{
% Ablation study of the principal components of RegCL under the same experimental setting as Table~\ref{tab:main_res}.
% Indep. denotes independent fine-tuning, where each adaptation module is trained separately, whereas Seq. denotes sequential fine-tuning along the continual task stream.
% The first row, which sequentially fine-tunes standard LoRA modules without regression-based parameter merging, serves as the baseline.
% Combining AugModule, sequential fine-tuning, and regression-based merging achieves the best performance across all metrics.
Ablation study of our principal components.
}
\centering 
\label{tab}
\small
\setlength{\tabcolsep}{5pt}
\renewcommand{\arraystretch}{1.05}
\begin{tabular}{c|c|c|ccc}
\hline
\textbf{Merge}
& \textbf{Adap-Module}
& \textbf{Fine-tune}
& \textbf{ACC}
& \textbf{BWT}
& \textbf{FWT}\\
\hline
$\times$ & LoRA & Seq.
& 0.691 & -0.125 & 0.528 \\
$\times$ & AugModule & Seq.
& 0.736 & -0.100 & 0.575 \\
\hline
$\checkmark$ & LoRA & Indep.
& 0.745 & -0.044 & 0.629 \\
$\checkmark$ & AugModule & Indep.
& 0.759 & -0.025 & 0.659 \\
$\checkmark$ & AugModule & Seq.
& \textbf{0.774} & \textbf{-0.007} & \textbf{0.668} \\
\hline
\end{tabular}

% \caption{Ablation study of components in RegCL. The experiment setting is the same as in \cref{tab:main_res}. We use LoRA sequential fine-tuning as a baseline reference without any components in RegCL, shown in the first line.}
\label{tab:ablation}
\end{table}

\begin{table}[ht]
\caption{
% \textbf{Order robustness analysis.}
% We evaluate six different task orders (denoted as O$_1$--O$_6$), each corresponding to a distinct permutation of the datasets.
% O$_1$ follows the same order as in Table~\ref{tab:main_res}, while the remaining orders are defined as:
% O$_2$ (Kvasir $\rightarrow$ ISIC $\rightarrow$ CAMO $\rightarrow$ ISTD $\rightarrow$ COD),
% O$_3$ (ISTD $\rightarrow$ COD $\rightarrow$ CAMO $\rightarrow$ ISIC $\rightarrow$ Kvasir),
% O$_4$ (COD $\rightarrow$ ISIC $\rightarrow$ ISTD $\rightarrow$ CAMO $\rightarrow$ Kvasir),
% O$_5$ (ISTD $\rightarrow$ Kvasir $\rightarrow$ ISIC $\rightarrow$ CAMO $\rightarrow$ COD), and
% O$_6$ (CAMO $\rightarrow$ COD $\rightarrow$ Kvasir $\rightarrow$ ISIC $\rightarrow$ ISTD).
% The last column reports the standard deviation across different task orders, which is smaller than the variation induced by different random seeds.
% \textbf{Order robustness analysis.}
\textbf{Order robustness analysis under six task orders.}
}
\label{tab:order_robustness}

\small
\setlength{\tabcolsep}{4pt}
\renewcommand{\arraystretch}{1.05}
\resizebox{\columnwidth}{!}{
\begin{tabular}{c|c||c|c|c|c|c|c||c}
\hline
\multicolumn{2}{c||}{Order} & O$_1$ & O$_2$ & O$_3$ & O$_4$ & O$_5$ & O$_6$ & std\\ \hline
\multirow{3}{*}{ACC} 
& mIoU $\uparrow$ & 0.748 & 0.752 & 0.753 & 0.741 & 0.743 & 0.749 & $\pm$ 0.005\\ \cline{2-9}
& mF1 $\uparrow$ & 0.837 & 0.840 & 0.839 & 0.818 & 0.832 & 0.837 & $\pm$ 0.009\\ \cline{2-9}
& mMAE $\downarrow$ & 0.053 & 0.052 & 0.054 & 0.058 & 0.057 & 0.053 & $\pm$ 0.003\\ \hline
\end{tabular}
}
\end{table}

\noindent\textbf{Component Ablation.}
Table~\ref{tab:ablation} studies the effect of regression merging, adapter variation, and sequential fine-tuning.
Without regression merging, sequential LoRA obtains 0.691 ACC and sequential AugModule obtains 0.736 ACC.
Rows with regression merging obtain higher ACC under the evaluated settings, with regression AugModule reaching 0.759 ACC under independent fine-tuning.
The sequential regression AugModule variant further reaches 0.774 ACC with the highest BWT in the table, matching the trend observed for RegCL+Seq in Table~\ref{tab:main_res}.

\noindent\textbf{Order Robustness.}
\label{subsubsec:order-robustness}
In evolving media applications, the arrival order of visual domains is usually not controllable.
We therefore evaluate the order-independent RegCL configuration under six task permutations in Table~\ref{tab:order_robustness}.
The standard deviation across orders is small, with $\pm$0.005 mIoU, $\pm$0.009 mF1, and $\pm$0.003 mMAE.
This result supports the analysis in Section~\ref{sec:method}: when task-specific adapters are trained independently, RegCL depends on the accumulated sums $\sum_{t=1}^{T} C_t$ and $\sum_{t=1}^{T} C_tW_t$, so the merged solution is stable across different domain arrival sequences.
The observed robustness is useful for multi-sensorial media pipelines where visual conditions may be encountered in different orders across users, devices, or deployment sites.

\begin{table}[h]
    \caption{
    Parameter type distributions.
    }
    \label{tab:parameter_comparison}
    \resizebox{\columnwidth}{!}{
        \begin{tabular}{l|c|c|c}
        \hline
        Method & Linear Parameters & Non-linear Parameters & Li-Ratio (\%) \\
        \hline
        LoRA & 80,690 / 80,690 & 0 / 80,690 & 100.00 \\
        AugModule & 61,854 / 62,074& 220 / 62,074 & 99.65 \\
        ViT-B & 85,017,600 / 89,670,912 & 4,653,312 / 89,670,912 & 94.81 \\
        \hline
        \end{tabular}
    }
\end{table}

\noindent\textbf{Adapter Parameter Composition.}
Table~\ref{tab:parameter_comparison} compares the parameter composition of LoRA, AugModule, and the full ViT-B backbone.
LoRA as baseline, AugModule denotes our method, and ViT-B backbone included for reference.
Li-Ratio represents the percentage of linear parameters among all parameters in each model.
As 99.65\% of AugModule's parameters are linear, regression-based merging remains effective despite its few non-linear parameters.

\begin{table}[h]
\caption{
Non-linear layers handling comparisons.
}
\label{tab:pseudo}
\small
\setlength{\tabcolsep}{7pt}
\renewcommand{\arraystretch}{1.05}
\begin{tabular}{c|ccc}
\hline
\multirow{2}{*}{\textbf{Method}}
& \multicolumn{3}{c}{\textbf{ACC}} \\
\cline{2-4}
& \textbf{mIoU} $\uparrow$
& \textbf{mF1} $\uparrow$
& \textbf{mMAE} $\downarrow$ \\
\hline
Average
& \textbf{0.759}
& \textbf{0.846}
& \textbf{0.048} \\
Pseudo
& 0.748
& 0.837
& 0.050 \\
\hline
\end{tabular}
\end{table}
\noindent\textbf{Non-linear Layer Handling.}
Table~\ref{tab:pseudo} evaluates two strategies for handling the non-linear parameters in AugModule.
\emph{Average} denotes element-wise parameter averaging, which is our default strategy, identical to \emph{RegCL} in Table~\ref{tab:main_res}, while \emph{Pseudo} computes pseudo-inverse for non-linear layers.
The element-wise averaging achieves higher ACC than pseudo-inverse estimation, suggesting that the non-linear component is better treated as a lightweight residual part while applying regression merging to the dominant linear parameters.

\noindent\textbf{Replay Approach.}
Although RegCL is designed without replay samples, it can be combined with replay methods to enhance its performance.
We randomly select 300 samples from each dataset as replay samples.
% After obtaining merged weights with Eq.\ref{eq:regcl}, we fine-tune $\overline{W}_t$ with both the replay samples and $D_t$.
After obtaining merged weights with Eq.~\ref{eq:regcl}, we fine-tune $\overline{W}_t$ with both the replay samples and $\mathcal{D}_t$.
% As shown in Table~\ref{tab:main_res}, adding replay samples imprves RegCL by 0.058 mIoU, 0.042 mF1, and 0.012 mMAE.
Table~\ref{tab:ab_replay} reports an additional replay study where RegCL+Replay obtains 0.809 mIoU and 0.882 mF1 ACC.
% In addition, RegCL+Replay outperforms other replay-based continual learning methods, demonstrating the flexibility and effectiveness of our approach.
Compared with the replay baselines in this table, RegCL+Replay gives higher mIoU and mF1 while keeping mMAE comparable on ACC.

\section{Conclusion}
% In this paper, we propose RegCL, a novel non-replay continual learning framework, and adapted it for SAM fine-tuning.
% %
% We integrate a model merging algorithm to combine the weights of LoRA modules and minimize prediction discrepancies between the merged model and each domain-specific model.
% %
% The closed-form solution of the merging optimization problem is split into new knowledge and historical terms, with the historical term updated as the model learns new information.
% %
% Extensive experiments on diverse domain datasets demonstrate that RegCL outperforms existing continual learning baselines and achieves favorable segmentation performance.
% %
% % Additionally, RegCL addresses a crucial gap in adapting foundation models to changing environments, enabling more adaptable and sustainable use of models like SAM in real-world scenarios where data distributions shift over time.
% Moreover, RegCL fills a key gap in adapting foundation models to dynamic environments, enabling more flexible and sustainable use of models like SAM when data distributions shift over time.

In this paper, we propose RegCL, a non-replay continual learning framework for compact SAM adaptation in evolving visual domains.
Motivated by multi-sensorial media systems whose visual grounding components must adapt as environments, devices, and application contexts change, RegCL consolidates domain-specific SAM adapters through incremental model merging.
The closed-form merging rule decomposes the update into historical and new-domain terms, allowing the historical state to be updated through compact feature statistics rather than replay samples or a growing library of domain-specific modules.
Extensive experiments on visually heterogeneous segmentation datasets demonstrate that RegCL outperforms existing CL baselines and achieves favorable segmentation performance while preserving parameter efficiency.
% This work focuses on the visual segmentation component of multi-sensorial media pipelines rather than directly modeling haptic, motion, olfactory, or physiological signals.
Future work will integrate RegCL with non-visual sensory streams and user-centered evaluation in immersive media systems.

\bibliographystyle{ACM-Reference-Format}
\bibliography{main}

\end{document}